\documentclass{article}

     \PassOptionsToPackage{sort,numbers}{natbib}

\usepackage[final]{neurips_2023}

\usepackage[colorlinks=true,breaklinks=true,bookmarks=false,citecolor=green]{hyperref}       
\usepackage[utf8]{inputenc} 
\usepackage[T1]{fontenc}    
\usepackage{url}            
\usepackage{booktabs}       
\usepackage{amsfonts}       
\usepackage{amssymb}     
\usepackage{amsmath}     
\usepackage{nicefrac}       
\usepackage{microtype}      
\usepackage{xcolor}         
\usepackage{bm}     
\usepackage{xfrac} 
\usepackage{array}
\usepackage{makecell}
\usepackage{multirow}
\usepackage{wrapfig}
\usepackage{enumitem}
\usepackage{float}

\usepackage{comment}  
\usepackage[ruled,vlined]{algorithm2e}	
\usepackage{lineno}
\usepackage{soul}

\newcommand{\PreserveBackslash}[1]{\let\temp=\\#1\let\\=\temp}
\newcolumntype{C}[1]{>{\PreserveBackslash\centering}p{#1}}
\newcolumntype{R}[1]{>{\PreserveBackslash\raggedleft}p{#1}}
\newcolumntype{L}[1]{>{\PreserveBackslash\raggedright}p{#1}}

\title{Human-Guided Shade Artifact Suppression in CBCT-to-MDCT Translation via Schrödinger Bridge with Conditional Diffusion}

\author{%
	Sung Ho Kang\textsuperscript{1} ~~ Hyun-Cheol Park\textsuperscript{2*} ~~ \\
	\textsuperscript{1}National Institute for Mathematical Sciences, Yuseong-gu, Daejeon 34047, Republic of Korea \\
	\textsuperscript{2}Department of Computer Engineering, Korea National University of Transportation,\\
	Chungju-si, Chungcheongbuk-do 27469, Republic of Korea \\
	\textsuperscript{*}Corresponding author: \texttt{hc.park@ut.ac.kr}
}

\begin{document}


\maketitle

\begin{abstract}
	We present a novel framework for CBCT-to-MDCT translation, grounded in the Schrödinger Bridge (SB) formulation, which integrates GAN-derived priors with human-guided conditional diffusion. Unlike conventional GANs or diffusion models, our approach explicitly enforces boundary consistency between CBCT inputs and pseudo targets, ensuring both anatomical fidelity and perceptual controllability. Binary human feedback is incorporated via classifier-free guidance (CFG), effectively steering the generative process toward clinically preferred outcomes. Through iterative refinement and tournament-based preference selection, the model internalizes human preferences without relying on a reward model. Subtraction image visualizations reveal that the proposed method selectively attenuates shade artifacts in key anatomical regions while preserving fine structural detail. Quantitative evaluations further demonstrate superior performance across RMSE, SSIM, LPIPS, and Dice metrics on clinical datasets—outperforming prior GAN- and fine-tuning-based feedback methods—while requiring only 10 sampling steps. These findings underscore the effectiveness and efficiency of our framework for real-time, preference-aligned medical image translation.
\end{abstract}

\begin{figure}[H]
	\centering
	\includegraphics[width=\textwidth]{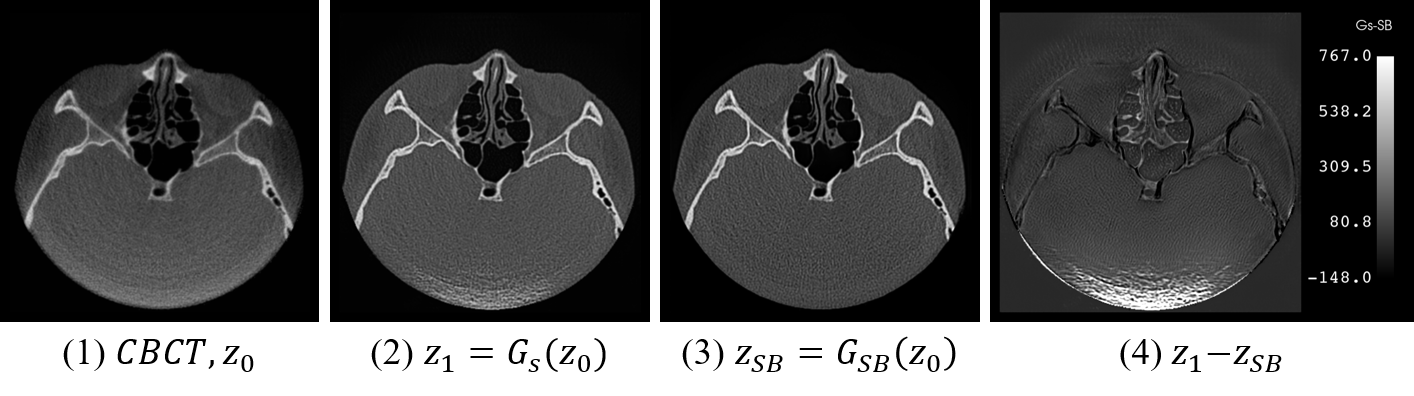}
	\caption{
		Representative example demonstrating effective shade artifact suppression guided by human feedback. From left to right: (1) input CBCT image; (2) output from the pretrained generator \(G_s\); (3) output from our SB-based generator \(G_{\mathrm{SB}}\); and (4) difference map between \(G_s\) and \(G_{\mathrm{SB}}\). The highlighted differences in soft-tissue regions—particularly in the posterior area—indicate that our model has successfully learned to reduce artifacts while preserving anatomical structures. This serves as a motivating example for the proposed preference-guided framework.
	}
	\label{fig:intro-diffmap}
\end{figure}

\section{Introduction}

Medical image translation, particularly the conversion from Cone-Beam Computed Tomography (CBCT) to Multi-Detector Computed Tomography (MDCT), has progressed rapidly in recent years with the advent of generative models. Among these, Generative Adversarial Networks (GANs) have demonstrated notable success in preserving anatomical fidelity in image-to-image translation tasks within medical imaging~\cite{park2022unpaired}. However, GANs inherently suffer from the quality–sampling–diversity trilemma, whereby enhancing one aspect frequently compromises the others. This trade-off often results in outputs that are biased toward specific modes and accompanied by unintended artifacts~\cite{xiao2021tackling}. Furthermore, even fine-tuned GANs may not overcome the limitations of pretrained models; they tend to overfit specific datasets, thereby reducing generalizability and introducing additional artifacts when applied to unseen data~\cite{park2022unpaired}.

As an illustrative motivation, Figure~\ref{fig:intro-diffmap} shows a representative case where our method significantly suppresses shade artifacts through human preference guidance. Starting from a CBCT input that exhibits pronounced artifacts, the pretrained generator \(G_s\) fails to remove them. In contrast, our SB-based generator \(G_{\mathrm{SB}}\), trained with binary human feedback, effectively attenuates these artifacts. The final subtraction image between \(G_s\) and \(G_{\mathrm{SB}}\) highlights the localized intensity changes—particularly in soft-tissue regions—where artifact suppression has occurred. This visual comparison underscores the model’s ability to learn a semantically meaningful correction trajectory from noisy, artifact-prone inputs. To address these challenges, recent studies have explored human-in-the-loop strategies. One approach involves manually filtering overfitted image pairs and reconstructing a new dataset under a paired training scheme to mitigate overfitting~\cite{park2022unpaired}. Alternatively, unpaired training schemes have incorporated human feedback and style transfer to fine-tune shade artifacts while preserving training data and minimizing model complexity~\cite{park2025shade}. Nevertheless, the requirement to train a separate reward model for incorporating human feedback introduces additional architectural complexity and computational burden, thereby posing challenges for clinical scalability and practical deployment.

In this context, Direct Preference Optimization (DPO)~\cite{rafailov2023direct}, which has recently gained attention in the language modeling field (e.g., ChatGPT), offers a promising alternative. DPO enables direct optimization using human preference data without the need for an explicit reward model, thereby simplifying training and reducing potential bias~\cite{rafailov2023direct}.

Diffusion models are increasingly being applied to medical imaging tasks due to their strong capacity for generating diverse outputs. These models have demonstrated high performance in various applications—including denoising, super-resolution, and domain adaptation—and have proven effective in producing high-quality, expressive medical images~\cite{ho2020denoising, sohl2015deep, song2020score}.

Of particular interest is the Image-to-Image Schrödinger Bridge (I\textsuperscript{2}SB), a nonlinear diffusion-based framework that directly connects the probabilistic distributions between image pairs. Its interpretable and efficient generation structure effectively controls probability flow, mitigates overfitting, and adapts flexibly to diverse scenarios~\cite{liu20232}.

\begin{figure}[t]
	\centering
	\includegraphics[width=\linewidth]{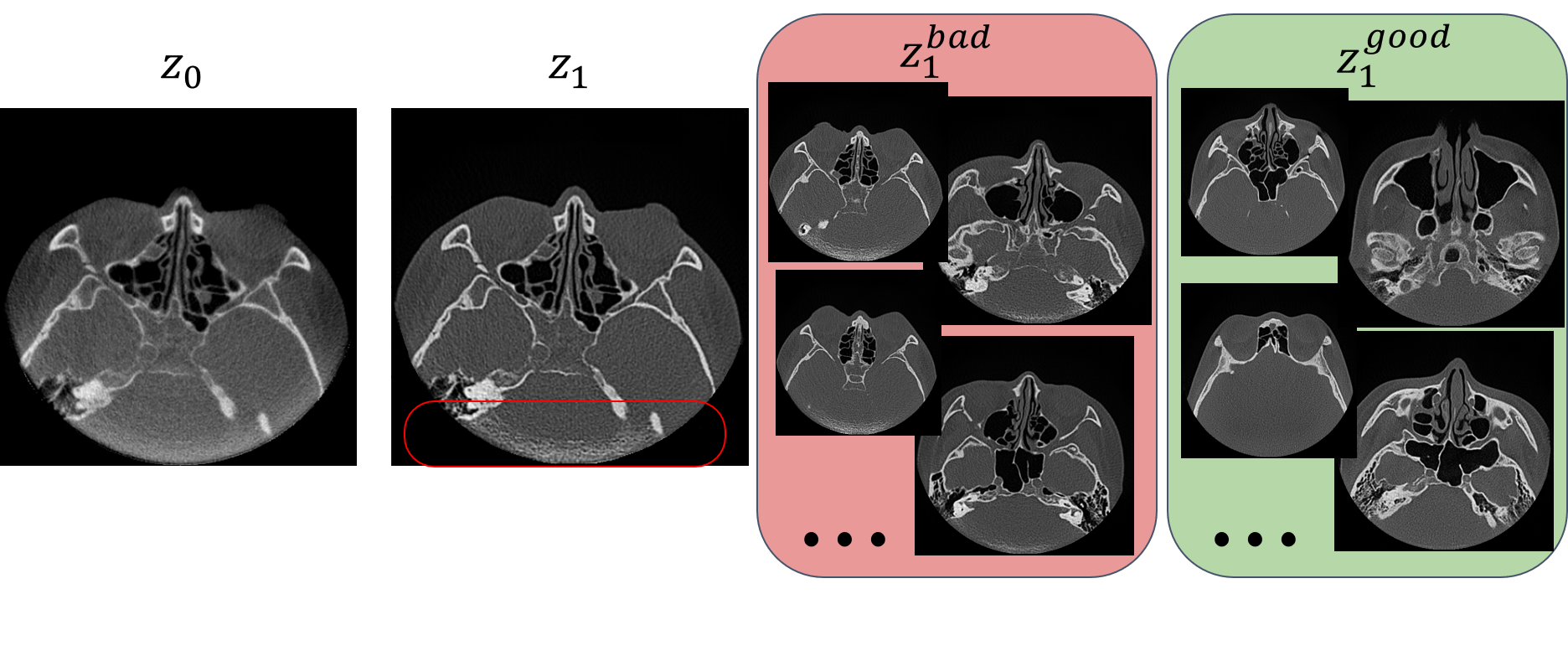}
	\caption{
		Visualization of input CBCT image $\bm{z}_0$ and its pseudo-target counterpart $\bm{z}_1 = G_s(\bm{z}_0)$ generated by an unpaired CycleGAN model. The second image highlights a region (red circle) with severe shade artifact—a common failure mode of GANs due to mode collapse. Notably, the ``bad'' sample $\bm{z}_1^{\text{bad}}$ exhibits a prominent shade artifact in the occipital region, which compromises anatomical fidelity. To address this, we collect human feedback on $\bm{z}_1$ samples and categorize them into ``bad'' (left) and ``good'' (right) groups based on clinical quality and artifact severity. These preference annotations are later used as binary reward signals $r\in\{0, 1\}$ to guide conditional generation.
	}
	\label{fig:z0-z1-feedback}
\end{figure}

To illustrate the core challenge addressed in this work, Figure~\ref{fig:z0-z1-feedback} shows an example of the CBCT input $\bm{z}_0$ and its translated pseudo-target $\bm{z}_1 = G_s(\bm{z}_0)$ produced by a pretrained CycleGAN~\cite{park2022unpaired}. A pretrained CycleGAN generator \( G_s \), trained in an unpaired fashion, is used to generate MDCT-like outputs from CBCT inputs, aiming to reduce artifacts while preserving anatomical structures. While such unpaired models can synthesize anatomically plausible structures, they often exhibit shade artifacts and inconsistencies due to mode collapse. In particular, we highlight (with a red circle) a typical failure case where structural fidelity is compromised. To mitigate these effects, we manually curate the $\bm{z}_1$ outputs into ``good'' and ``bad'' sets through human evaluation. These preference labels are then converted into binary feedback values $r\in\{0, 1\}$, which are incorporated into our conditional generative model to steer the reconstruction path toward clinically desirable outcomes.

Inspired by DPO~\cite{rafailov2023direct}, which enables human-aligned generation without the need for an explicit reward model, we aim to translate a similar philosophy into the diffusion domain. Rather than relying on a separate reward network to model preference, we introduce a binary feedback signal $r\in\{0, 1\}$ that conditions the diffusion process via Classifier-Free Guidance (CFG)~\cite{ho2022classifier}. This allows the model to flexibly explore diverse generation paths and converge toward human-desired outcomes. In essence, our approach leverages the expressiveness and diversity of diffusion while preserving the simplicity and alignment capabilities of DPO, thus providing a practical alternative for clinical applications where reward model training is prohibitive.

In this study, we propose a novel framework that leverages high-fidelity images generated by GANs as priors, integrated with the I\textsuperscript{2}SB framework and enhanced by human feedback. Our approach aims to capitalize on the interpretability and efficiency of I\textsuperscript{2}SB while complementing the strengths of both paired and unpaired training schemes. Specifically, we utilize the output of a GAN trained in an unpaired setting as a pseudo-target prior, enabling paired learning within the I\textsuperscript{2}SB framework. This design addresses the quality–sampling–diversity trilemma by leveraging the diversity of diffusion models, while guiding the generation trajectory using CFG~\cite{ho2022classifier} and refining outputs through human feedback. Furthermore, we incorporate incremental learning via class injection based on preference annotations, allowing explicit control over clinically desirable image attributes. As a result, our method improves the accuracy of CBCT-to-MDCT translation and effectively suppresses artifacts, including those arising from mode collapse.

\section{Related Work}

\subsection{GANs in Medical Imaging}
GANs have proven effective for medical image translation tasks such as denoising~\cite{yang2018low}, super-resolution~\cite{de2021impact}, and cross-modality synthesis~\cite{hiasa2018cross}, owing to their capacity for learning complex data distributions and preserving anatomical fidelity~\cite{park2022unpaired}. However, challenges including mode collapse, unstable training dynamics, and residual artifacts such as shading and hallucinations remain problematic~\cite{arjovsky2017wasserstein,wolterink2017generative,thanh2020catastrophic}. To mitigate these issues, \cite{park2025shade} proposed a fine-tuning approach that incorporates human feedback and style transfer, leveraging a reward model to classify artifact-laden outputs and selectively guide training.

\subsection{Diffusion Models for Medical Image Generation}
Diffusion probabilistic models such as DDPMs~\cite{ho2020denoising} and SGMs~\cite{song2020score} offer improved training stability and sampling diversity over GANs. They have shown strong performance in medical imaging applications including denoising, domain adaptation, and inpainting~\cite{chung2022come, wolleb2022diffusion}, particularly under noisy or low-dose acquisition settings. More recently, Li et al.~\cite{li2023zero} proposed a Frequency-Guided Diffusion Model (FGDM) for zero-shot CBCT-to-CT translation, which utilizes high-frequency priors from CBCTs to guide the generation of anatomically faithful CT-like images, without requiring any paired data for training. While FGDM is fully automated and performs robustly across domains, it lacks user-controllable mechanisms for preference-driven refinement, which may limit its clinical flexibility compared to human-guided frameworks like ours.

\subsection{Hybrid GAN–Diffusion Models}
Hybrid models combining GANs and diffusion have been proposed to exploit the strengths of both paradigms~\cite{xiao2021tackling,bond2022unleashing,ozbey2023unsupervised,zhao2024diffgan, chen2024hidiff}. For example, SynDiff~\cite{ozbey2023unsupervised} employs conditional diffusion with an adversarial projector to refine outputs. However, these models often require complex architectures and careful loss balancing, which may hinder robustness and scalability.

\subsection{Schrödinger Bridge Models}
Schrödinger Bridge (SB) models offer a probabilistic framework for optimal transport between source and target distributions, enabling more interpretable and structure-aware generation than traditional diffusion approaches. The DSB framework~\cite{de2021diffusion} formalized this connection by interpreting score-based models as a special case of SB, improving convergence with fewer steps via iterative proportional fitting. I\textsuperscript{2}SB~\cite{liu20232, li2024diffusion, zhou2024cascaded} advanced this idea by eliminating simulation during training and achieving efficient generation, making SB models practical for high-dimensional medical image translation. Building on this, we incorporate human feedback as a conditioning signal in SB-based sampling, using GAN-derived pseudo-targets as priors. This approach addresses limitations such as mode collapse and semantic drift, while improving cross-modality consistency and anatomical fidelity.

\subsection{Human Preference Alignment}
Traditional integration of human feedback into generative models relies on reinforcement learning with human feedback ~\cite{christiano2017deep, stiennon2020learning}, which requires a separate reward model and extensive annotations. Such frameworks are complex, error-prone, and ill-suited to medical domains where feedback is costly and interpretability critical. As an alternative, DPO~\cite{rafailov2023direct} enables reward-model-free training by directly optimizing from preference data using a contrastive objective. While initially developed for language models, its philosophy aligns well with clinical image translation, where model alignment and transparency are essential. In this work, we extend DPO principles into the diffusion setting by injecting binary feedback $r\in\{0, 1\}$ via CFG, enabling soft control of image semantics without relying on a learned reward model. This improves scalability and controllability, offering a more practical solution for preference alignment in safety-critical applications.

\begin{figure}[t]
	\centering
	\includegraphics[width=\textwidth]{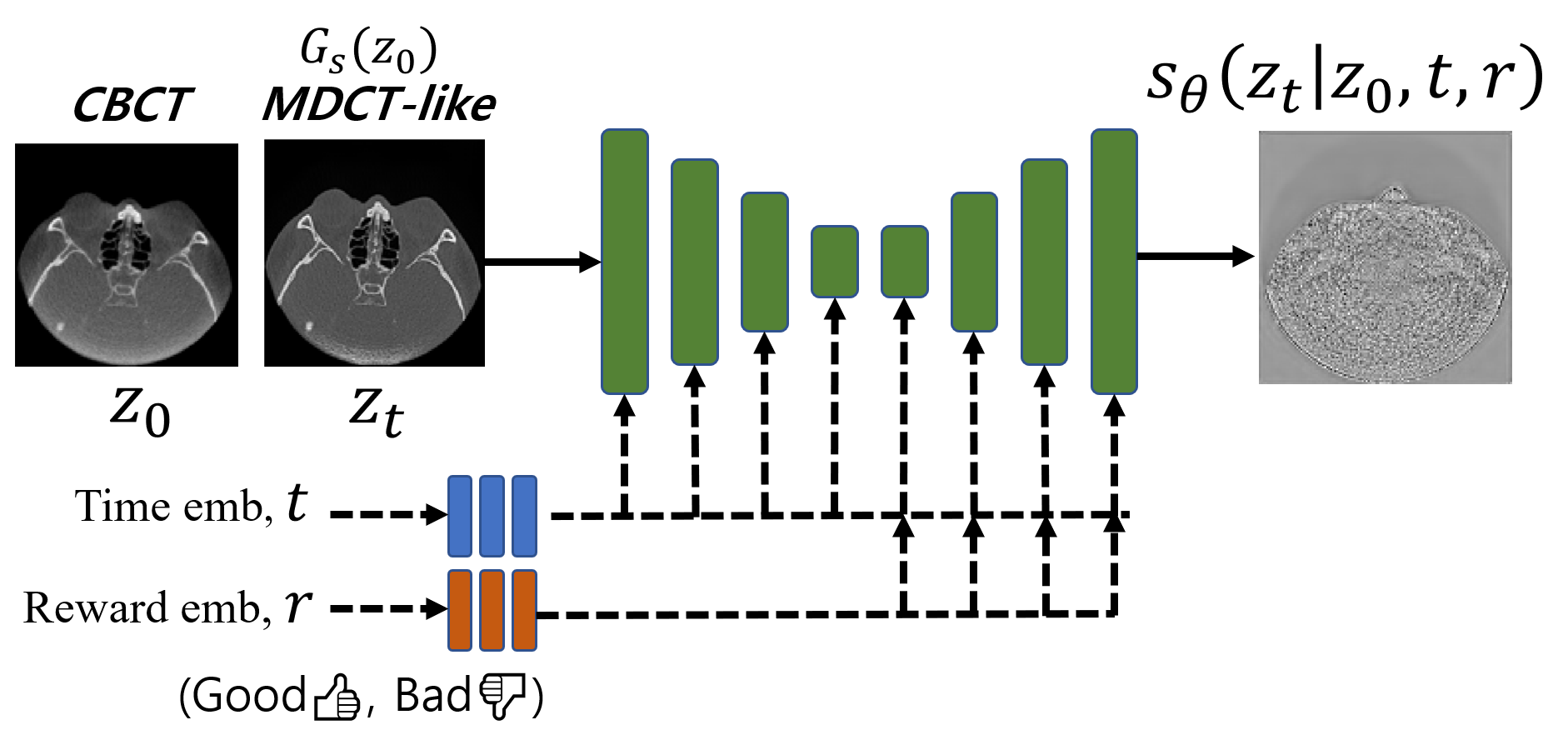}
	\caption{
		Overall architecture of the proposed framework. The input CBCT image $\bm{z}_0$ is translated into a pseudo-target image $\bm{z}_1 = G_s(\bm{z}_0)$ via a pretrained generator. The SB formulation then defines a stochastic path between $\bm{z}_0$ and $\bm{z}_1$, with intermediate latent states $\bm{z}_t$ denoised by a reward-conditioned score network $s_\theta(\bm{z}_t \mid \bm{z}_0, t, r)$. The UNet takes as input the noisy sample $\bm{z}_t$, the source $\bm{z}_0$, and two conditioning embeddings: timestep $t$ and binary human feedback $r\in\{0, 1\}$, indicating expert preference (0 = Good, 1 = Bad). To guide the sampling process, the time embedding $\gamma(t)$ is added as a residual bias to both the encoder and decoder layers to maintain diffusion dynamics, while the reward embedding $\gamma(r)$ is multiplicatively applied to the decoder layers only to convey semantic preferences. This reward-conditioned decoding enables the network to adaptively modulate generation according to both anatomical fidelity and clinical desirability.
	}
	\label{fig:network-architecture}
\end{figure}

\section{Methodology}\label{sec:method}
In this study, we propose a \textit{Conditional Schr\"{o}dinger Bridge} framework tailored for CBCT-to-MDCT translation that aims to reduce shade artifacts by integrating human feedback loops. Inspired by the I\textsuperscript{2}SB~\cite{liu20232}, our diffusion model explicitly accounts for two boundary states: the initial CBCT distribution (\(\bm{z}_0\)) and the pseudo-target distribution (\(\bm{z}_1\)), the latter being derived from a high-fidelity GAN prior. Unlike standard diffusion methods that rely solely on noise-adding and noise-removal processes via stochastic differential equations (SDEs), this SB-based approach enforces boundary consistency and thus enhances interpretability and stability in the generative path. Additionally, we incorporate CFG~\cite{ho2022classifier} and incremental learning strategies based on human preference data, allowing the model to refine outputs according to clinical or expert feedback without the need for a separate reward model. Consequently, our framework addresses the quality–sampling–diversity trilemma often encountered in GAN-based methods~\cite{xiao2021tackling, park2022unpaired}, while preserving anatomical fidelity and minimizing artifact intrusion. By unifying the interpretability of SB-driven diffusion, the flexibility of GAN priors, and the efficiency of direct human feedback loops, the proposed method offers a principled and efficient solution for CBCT-to-MDCT translation in real-world clinical settings.

\subsection{Schrödinger Bridge Formulation: Forward and Reverse SDEs}

The SB framework formulates a stochastic process that transports probability mass between two 
empirical boundary distributions---the source distribution $p_0$ supported by samples $\bm z_0$ and the target distribution $p_1$ supported by samples $\bm z_1$.  
Unlike conventional denoising diffusion probabilistic models~(DDPMs), which begin their generation from an 
isotropic Gaussian prior, the SB formulation constructs an optimal probabilistic path connecting the two empirical distributions.  
This bidirectional structure has been shown to improve both the stability and the interpretability of medical-image translation tasks~\cite{liu20232}.  

Throughout this paper we adopt the standard SB assumption that the deterministic drift is omitted, i.e. $\bm f_t\equiv \bm 0$, and the dynamics are driven purely by a time–dependent Brownian diffusion coefficient $\beta_t>0$.  

\subsubsection{Forward SDE}
The forward process gradually perturbs the source state $\bm z_0$ toward the target distribution and is given by
\begin{equation}
	d\bm z_t = \sqrt{\beta_t}\,d\bm w_t, \qquad t\in[0,1],
	\label{eq:forward_sde}
\end{equation}
where $\bm w_t$ denotes a standard Brownian motion in $\mathbb R^{d}$.  
Because the process starts from $\bm z_0$ and terminates at $\bm z_1$, the injected noise level $\beta_t$ is scheduled symmetrically with respect to $t=\tfrac12$ so that the marginal variance first increases and then decreases, yielding a smooth, bidirectional bridge rather than a one–way diffusion to Gaussian noise.

\subsubsection{Reverse SDE}
The reverse‑time dynamics transport the state from the target distribution back to the source distribution and read
\begin{equation}
	d\bm z_t = -\beta_t \, \nabla_{\!\bm z_t}\log \hat\Psi(\bm z_t,t)\,dt + \sqrt{\beta_t}\,d\bar{\bm w}_t,  \label{eq:reverse_sde}
\end{equation}
where $\hat\Psi(\bm z_t,t)$ is the dual potential that, together with the forward potential $\Psi$, solves the coupled Schrödinger system of partial differential equations~\cite{chen2021stochastic}.  
Note the sign reversal in the drift term compared to the forward SDE and the introduction of $\hat\Psi$: this choice is essential for the SB formulation and guarantees that the time‑reversed process in Eq.~\eqref{eq:reverse_sde} shares the same path measure as Eq.~\eqref{eq:forward_sde}.

\subsubsection{Closed‑form Gaussian Intermediates}
When both endpoints $\bm z_0$ and $\bm z_1$ are given, every intermediate latent variable $\bm z_t$ has a closed‑form Gaussian distribution
\begin{align}
	\bm \mu_t &= \frac{\bar\sigma_t^{2}}{\bar\sigma_t^{2}+\sigma_t^{2}}\,\bm z_0 + \frac{\sigma_t^{2}}{\bar\sigma_t^{2}+\sigma_t^{2}}\,\bm z_1, \label{eq:mu_t}\\[4pt]
	\bm \Sigma_t &= \frac{\sigma_t^{2}\,\bar\sigma_t^{2}}{\bar\sigma_t^{2}+\sigma_t^{2}}\,\bm I,  \label{eq:sigma_t}
\end{align}
where $\sigma_t^{2}=\int_{0}^{t}\!\beta_\tau\,d\tau$ and $\bar\sigma_t^{2}=\int_{t}^{1}\!\beta_\tau\,d\tau$.  
These expressions mirror the posterior of DDPMs, but here they arise from the bidirectional bridge and remain symmetric in $\bm z_0$ and $\bm z_1$.

In our CBCT–to-MDCT translation task, the target sample $\bm z_1$ is obtained by a CycleGAN‑based generator $G_s$ trained on unpaired data~\cite{park2022unpaired}.  We therefore set $\bm z_1 = G_s(\bm z_0)$ and treat it as a pseudo‑target guiding the reverse dynamics.

\subsection{Parameterization and Objective}
To parameterize the reverse‑time score in Eq.~\eqref{eq:reverse_sde} we employ a condition‑aware network $s_{\theta}(\bm z_t\,|\,\bm z_0, t, r)$ that approximates $-\bm \Sigma_t^{-1}(\bm z_t-\bm \mu_t)$, with $r\!\in\{0,1\}$ representing the binary human preference signal newly introduced in this work.  
The corresponding score‑matching loss is
\begin{equation}
	\mathcal L_{\text{score}}(\theta) = \mathbb E_{\bm z_0,\bm z_1, t, \bm z_t}\!\Bigl[\bigl\|\,s_{\theta}(\bm z_t\,|\,\bm z_0,t,r) + \bm \Sigma_t^{-1}(\bm z_t-\bm \mu_t)\bigr\|^{2}\Bigr],
	\label{eq:score_loss_revised}
\end{equation}
where the expectations are taken exactly as in the original formulation.  

\subsubsection{Simplified Objective}
When computational efficiency outweighs exactness we adopt the DDPM‑style surrogate
\begin{equation}
	\mathcal L_{\text{naive}} = \bigl\|\,s_{\theta}(\bm z_t,t) - (\bm z_t-\bm z_1)/\sigma_t\bigr\|^{2},
	\label{eq:naive_loss_revised}
\end{equation}
which has proven effective in practice~\cite{ho2020denoising, dhariwal2021diffusion,liu20232}.

\subsection{Conditional Diffusion with Binary Feedback}
We further extend the diffusion model to incorporate a binary feedback signal $r\in\{0, 1\}$, representing human preference annotations. The intermediate latent variable \(\bm{z}_t\) is sampled from a Gaussian distribution using the closed-form expressions for mean and covariance as defined in Eq.~\ref{eq:mu_t} and Eq.~\ref{eq:sigma_t}. Specifically, at each timestep \(t\), the denoising score function is conditioned not only on the sampled noisy state \(\bm{z}_t\), but also on the original source CBCT image \(\bm{z}_0\), the diffusion timestep \(t\), and the binary feedback \(r\):
\begin{equation}
	\bm{s}_\theta(\bm{z}_t \mid \bm{z}_0,\, t,\, r) = \nabla_{\bm{z}_t} \log p_\theta(\bm{z}_t \mid \bm{z}_0,\, t,\, r).
\end{equation}

The conditional score \(\bm{s}_\theta\) is parameterized by a neural network. To practically implement this conditioning, we construct the input vector to the score network by concatenating feature embeddings:
\begin{equation}
	\bm{c}_t := [\bm{z}_t;\, \bm{z}_0;\, \gamma(t);\; \gamma(r)],
\end{equation}
where \(\gamma(t)\) is a positional embedding (e.g., sinusoidal), and \(\gamma(r)\) is a learnable embedding obtained via a neural projection of the binary feedback. We then pass \(\bm{c}_t\) through a \(\mathrm{UNet}_\theta\), yielding
\begin{equation}
	\bm{s}_\theta(\bm{z}_t \mid \bm{z}_0,\, t,\, r) = \mathrm{UNet}_\theta(\bm{c}_t).
\end{equation}

\subsubsection{Reward and Time Embedding Layers}
To integrate preference information more effectively, the binary feedback score \(r\) is projected into a high-dimensional embedding via a multi-layer perceptron (MLP), producing reward embedding in \(\mathbb{R}^{C\times1\times1}\). Similarly, the diffusion timestep \(t\) is embedded using a positional encoding(such as sinusoidal functions), followed by a MLP to produce time embedding in \(\mathbb{R}^{C \times 1 \times 1}\). The time embedding is applied at all layers to preserve diffusion dynamics, while the reward embedding is exclusively applied at the decoder layers to guide high-level semantic refinement. Both embeddings are computed once per step and used accordingly throughout the network.

At each decoder level, the reward embedding is applied multiplicatively to modulate high-level semantic preferences, while the time embedding is added as a residual bias to preserve the diffusion dynamics (see Figure~\ref{fig:network-architecture}). This decoder-only reward conditioning is motivated by the intuition that higher-level semantic refinements—such as artifact suppression or preference-driven enhancement—are best applied in later reconstruction phases. This approach is conceptually similar to ControlNet~\cite{zhang2023controlnet} and D3PO~\cite{borso2025d3po}, which localize structural or preference conditioning in targeted parts of the network to balance anatomical fidelity with semantic flexibility.

\subsection{CFG-Based Preference Alignment with Incremental Feedback Learning}
To steer the generative process toward human-preferred attributes, we adopt CFG~\cite{ho2022classifier}. Specifically, we linearly combine conditional and unconditional score predictions as:
\begin{equation}
	\tilde{\bm{s}}(\bm{z}_t) = (1 + w)\cdot \bm{s}_\theta(\bm{z}_t \mid \bm{z}_0,\, t,\, r) - w \cdot \bm{s}_\theta(\bm{z}_t \mid \varnothing),
\end{equation}
where \(w > 0\) is the guidance scale, and \(\bm{s}_\theta(\cdot \mid \varnothing)\) denotes the unconditional score obtained by nullifying the preference signal \(r\). By adjusting \(w\), we can amplify or diminish the influence of the feedback conditioning.

Beyond generating preferred samples, CFG also serves as a mechanism for exploratory refinement. Starting from a ``bad'' sample identified by human annotation, we generate alternative reconstructions with increasing guidance scales \(w \in \{w_1, \dots, w_n\}\), each corresponding to a different CFG strength. This process produces a candidate set \(\tilde{\mathcal{D}}\) of reconstructions from the same input CBCT image \(\bm{z}_0\), as detailed in Algorithm~\ref{alg:cfg-feedback}. To identify the most perceptually favored reconstruction, we apply a tournament-based selection framework to \(\tilde{\mathcal{D}}\) (Algorithm~\ref{alg:tournament}). All candidates are first sorted by patient ID and slice number to maintain anatomical consistency. For each slice, the reconstructions are compared in a sequence of 1:1 pairwise matchups, where expert raters iteratively choose the more preferred image $x_\text{win}$. Through successive elimination, a final winner is determined per slice, forming the human-curated dataset \(\mathcal{Z}_{\text{pref}}\). This preferred dataset is then used to augment the training set \(\mathcal{D}_{\text{train}}\) for incremental fine-tuning. This framework enables the model to gradually adapt to human perceptual preferences without requiring an explicit reward model. Additionally, the structured tournament process reduces intra-rater bias and improves feedback reliability.

\begin{figure}[t]
	\centering
	\includegraphics[width=\textwidth]{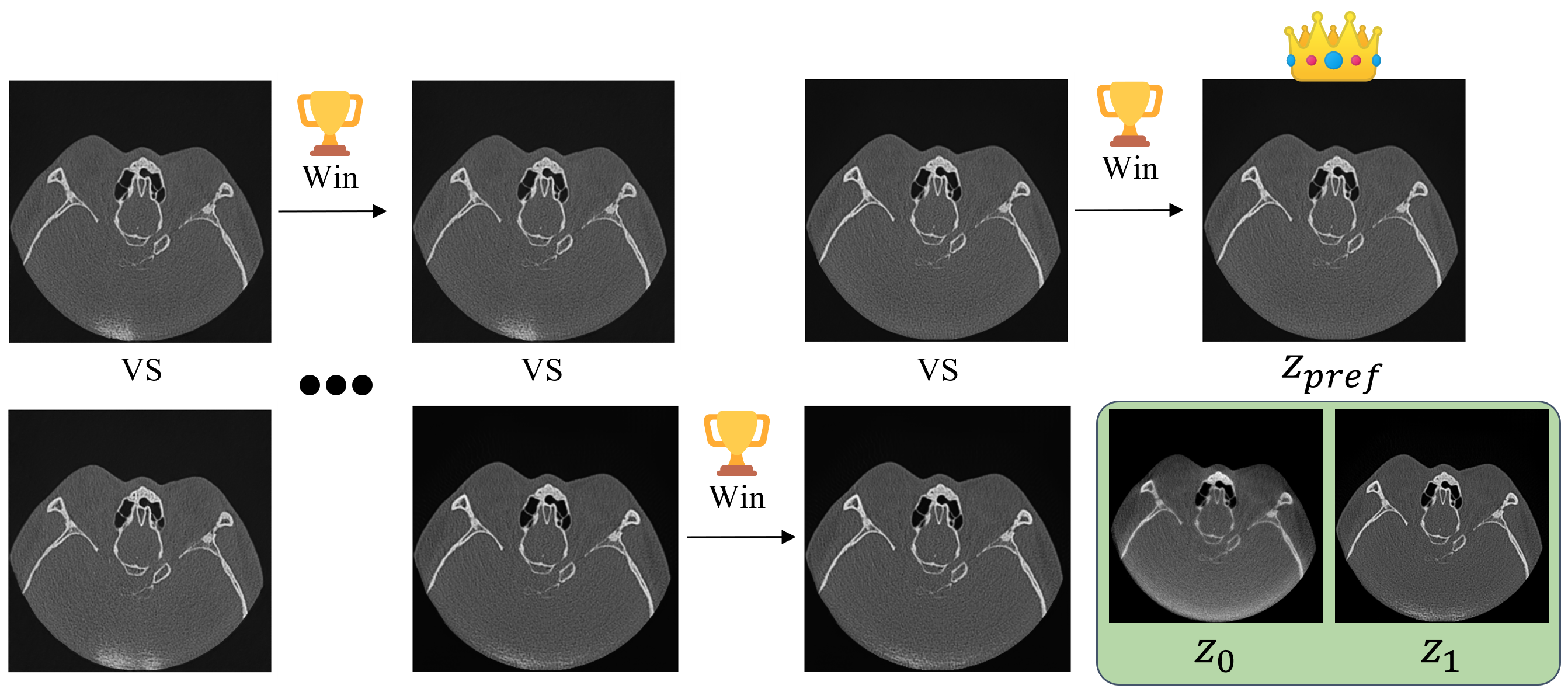}
	\caption{
		Visualization of CFG-guided sampling and tournament-based preference selection. Each column represents a candidate reconstruction generated under different combinations of guidance scale \(w_j\) and model checkpoint \(\theta_k\). In each column, human evaluators iteratively compare image pairs (marked with trophy icons) in a tournament-style process, ultimately selecting the most clinically plausible reconstruction (marked with a crown). This process enables preference-driven refinement of artifact suppression in CBCT-to-MDCT translation.
	}
	\label{fig:cfg-feedback-example}
\end{figure}

\begin{algorithm}[t]
	\caption{CFG-Guided Sampling and Incremental Feedback Learning}
	\label{alg:cfg-feedback}
	\SetKwInOut{Input}{Input}
	\SetKwInOut{Output}{Output}

	\Input{Initial CBCT image $\bm{z}_0$ with ``bad'' rating, pretrained model $s_\theta$, guidance scales $\{w_1, w_2, \dots, w_n\}$}
	\Output{Updated model after incremental feedback learning}

	Initialize temporary dataset $\tilde{\mathcal{D}} \leftarrow \emptyset$ \;

	\ForEach{$w \in \{w_1, \dots, w_n\}$}{
		\textit{/* Requesting improved output by setting $r = 0$ (i.e., ``good'') */} \\
		Set $r \leftarrow 0$ \;

		Sample $\bm{z}_t^{(w)}$ using CFG: $\tilde{\bm{s}}(\bm{z}_t^{(w)}) = (1 + w)\, s_\theta(\bm{z}_t^{(w)} \mid \bm{z}_0, t, r) - w \, s_\theta(\bm{z}_t^{(w)} \mid \varnothing)$ \;

		Sample output $\hat{\bm{x}}^{(w)}$ from $\bm{z}_t^{(w)}$ using a sampling step (e.g., DDIM) \;

		Add $(\bm{z}_0, \hat{\bm{x}}^{(w)})$ to $\tilde{\mathcal{D}}$ \;
	}

	Apply tournament selection (Algorithm~\ref{alg:tournament}) to $\tilde{\mathcal{D}}$ to obtain preferred set $\mathcal{Z}_{\text{pref}}$ \;

	Augment training set: $\mathcal{D}_{\text{train}} \leftarrow \mathcal{D}_{\text{train}} \cup \mathcal{Z}_{\text{pref}}$ \;

	Fine-tune $s_\theta$ on $\mathcal{D}_{\text{train}}$ \;
\end{algorithm}

\begin{algorithm}[t]
	\caption{Tournament-based Feedback Collection for Incremental Learning}
	\label{alg:tournament}
	\KwIn{Generated image set $\tilde{\mathcal{D}}$ per slice, sorted by patient ID and slice number; \\
		\quad Checkpoint set $\{\theta_k\}$; CFG scales $\{w_j\}$; Human evaluator $\mathcal{H}$}
	\KwOut{Preference-labeled image set $\{(x, r)\} \in \mathcal{Z}_{\text{pref}}$ with $r = 0$ (marked as ``good'') for fine-tuning}

	\ForEach{patient $p$}{
		\ForEach{slice $s$ of patient $p$}{
			Initialize candidate pool $\mathcal{C}_{p,s} \leftarrow$ \{images generated from all $(\theta_k, w_j)$\};

			\While{$|\mathcal{C}_{p,s}| > 1$}{
				Randomly sample image pairs $(x_i, x_j)$ from $\mathcal{C}_{p,s}$\;
				$x_\text{win} \leftarrow \mathcal{H}.compare(x_i, x_j)$\;
				$\mathcal{C}_{p,s} \leftarrow \mathcal{C}_{p,s} \cup \{x_\text{win}\} \setminus \{x_i, x_j\}$\;
			}

			Add final winner labeled as ``good'' ($r = 0$): $\mathcal{Z}_{\text{pref}} \leftarrow \mathcal{Z}_{\text{pref}} \cup \{(\mathcal{C}_{p,s}, 0)\}$\;
		}
	}

	\Return $\mathcal{Z}_{\text{pref}}$
\end{algorithm}

\section{Experiments}

\subsection{Dataset and Experiment Setup}

We used 20 CBCT and 28 MDCT volumes acquired under Institutional Review Board (IRB) approval. CBCT data were acquired using a circular-trajectory scanner (Xoran CAT, USA), and MDCT with a helical-trajectory scanner (SOMATOM Definition Flash, Siemens, Germany), both with in-plane resolutions around 0.4~mm.

Prior to training, MDCT volumes were preprocessed to align with the spatial resolution and anatomical framing of CBCT images. Each MDCT volume was first resampled to a voxel size of $0.40\,\text{mm} \times 0.40\,\text{mm}$ in the axial plane. Then, the images were cropped to a fixed size of $384 \times 384$ pixels centered at an anatomically defined point—approximately 2\,mm anterior to the Sella—using a landmark detection algorithm~\cite{kang2021landmark} to ensure consistent field-of-view alignment across subjects.

Among the 20 CBCT subjects, 17 were used for training and 3 for testing. From each CBCT image $z_0$, pseudo-targets $z_1 = G_s(z_0)$ were generated via a pretrained unpaired CycleGAN. Human experts evaluated shade artifacts and annotated 4,075 slices as ``good'' and 712 as ``bad'' based on clinical quality and artifact severity (see Figure~\ref{fig:z0-z1-feedback}). These feedback labels formed the basis for preference-conditioned training and evaluation.

\subsection{Evaluation Metrics}

We evaluated performance using Root Mean Square Error (RMSE), Structural Similarity Index Measure (SSIM), Learned Perceptual Image Patch Similarity (LPIPS), and Dice coefficient (DC). These metrics were computed using expert-approved images from prior work \cite{park2022unpaired, park2025shade} as pseudo-ground-truths. For shade artifact evaluation, we additionally report Artifact Reduction Rate (ARR) and Artifact Reduction Success Rate (ARSR), following \cite{park2025shade}.

\subsection{Preference-Guided Sampling and Feedback}

To enable human-guided generation, we applied CFG as detailed in Algorithm~\ref{alg:cfg-feedback}. Specifically, we used 9 model checkpoints (corresponding to early, mid, and late training phases) and 6 guidance scales $w \in \{1.0, 2.0, 4.0, 5.0, 8.0, 10.0\}$ to generate diverse reconstructions per input slice. For each axial slice, CFG-based reconstructions were generated across all $(\theta_k, w_j)$ combinations, and two expert raters conducted tournament-style comparisons (Algorithm~\ref{alg:tournament}) to identify the most clinically preferred outputs. This process, illustrated in Figure~\ref{fig:cfg-feedback-example}, produced a curated dataset $\mathcal{Z}_{\text{pref}}$ used to fine-tune the model.

Our implementation follows the design of I\textsuperscript{2}SB~\cite{liu20232}, employing a UNet-based score network and a symmetric noise scheduling strategy for the diffusion process, where the noise variance $\beta_t$ follows a symmetric schedule that increases to a maximum at $t = \frac{1}{2}$ and decreases toward both boundaries. Intermediate states $\bm{z}_t$ are analytically sampled from Gaussian posteriors conditioned on boundary states $(\bm{z}_0, \bm{z}_1)$, enabling stable generation and efficient sampling. All training and evaluation settings were kept consistent to ensure reproducibility.

\section{Results and Discussion}

In many clinical settings, acquiring pixel-aligned ground-truth MDCT images corresponding to CBCT inputs is impractical due to ethical and technical constraints. To address this, we adopt the evaluation strategy introduced by~\cite{park2022unpaired}, in which enhanced CBCT images—validated by expert radiologists—serve as de facto ground-truths for quantitative assessment. CBCT images are commonly used in dental applications for reconstructing 3D surface mesh models of anatomical structures such as bones. In that study, two board-certified radiologists with over 20 years of clinical experience evaluated the generated CBCT-to-MDCT translations across 20 sample datasets, including assessments of overall image quality, shade artifact reduction in the maxillofacial region, and the delineation of bone boundaries. Quantitative analysis, based on expert-annotated bone segmentations, revealed that the enhanced images substantially improved structural fidelity without introducing or omitting unintended features. These refined outputs are thus treated as ground-truths for evaluating the performance of new generative models.

Following this precedent, we use the expert-approved images from that work as target ground-truths to measure anatomical consistency, artifact suppression, and surface reconstruction accuracy in our results. Dice coefficients are computed using the segmentation labels annotated by the same expert radiologists, with a fixed threshold value of 800 applied to ensure consistency~\cite{park2025shade}. Additionally, we evaluate the generated images by measuring RMSE, SSIM, and LPIPS~\cite{zhang2018unreasonable}, with respect to the outputs of the pretrained generator \(G_s\), which serves as a baseline reference.

\subsection{Shade Artifact Suppression}
To quantitatively evaluate the effectiveness of our proposed method in suppressing shade artifacts, we compared the outputs \(\bm{z}_1^{\text{bad}}\) from the pretrained GAN-based generator \(G_s\) and \(\bm{z}_{\mathrm{SB}}^{\text{bad}}\) generated by our SB-based generator \(G_{\mathrm{SB}}\). Following the evaluation protocol proposed by~\cite{park2025shade}, we computed two key metrics: ARR and ARSR. ARR measures the relative reduction of artifact intensity in regions known to exhibit shading, whileARSR quantifies the proportion of cases in which shade artifacts are deemed completely suppressed based on predefined criteria. Together, these metrics provide a comprehensive assessment of both voxel-level and case-level improvements in shade artifact suppression without requiring ground-truth MDCT references.

We evaluated these metrics on both the training dataset (\(N=712\)) and the test dataset (\(N=99\)) to assess not only the artifact suppression performance but also the generalizability of the model to unseen data. The results, summarized in Table~\ref{tab:arr-arsr-results}, show that our method (\(G_{\mathrm{SB}}\)) consistently achieves the highest scores across all metrics and datasets.

Compared to the Park2025 method~\cite{park2025shade}, which already incorporates human feedback and fine-tuning, our model improves ARSR on the test set from 95.96\% to 96.23\%, and ARR from 96.86\% to 96.98\%. The improvements are even more significant when compared to Park2022~\cite{park2022unpaired}, highlighting the contribution of our SB-based sampling and human-guided refinement pipeline.

These results underscore the effectiveness of our SB-based sampling framework, which not only reduces shade artifacts more reliably but also generalizes better across diverse clinical cases due to its human-guided and preference-aware design.

While the above metrics provide quantitative evidence of artifact suppression, visual interpretation is also essential. It is important to note that, due to the nature of \(\bm{z}_1^{\text{bad}}\), which by definition contains prominent shade artifacts, lower perceptual similarity scores (i.e., higher LPIPS values) between \(\bm{z}_1^{\text{bad}}\) and \(\bm{z}_{\mathrm{SB}}^{\text{bad}}\) often reflect effective artifact suppression rather than degradation in fidelity. In contrast, when using \(\bm{z}_1^{\text{good}}\) for structural fidelity assessment, lower LPIPS values do correspond to higher visual consistency and are thus considered desirable. 

\begin{figure}[t]
	\centering

	\includegraphics[width=\textwidth]{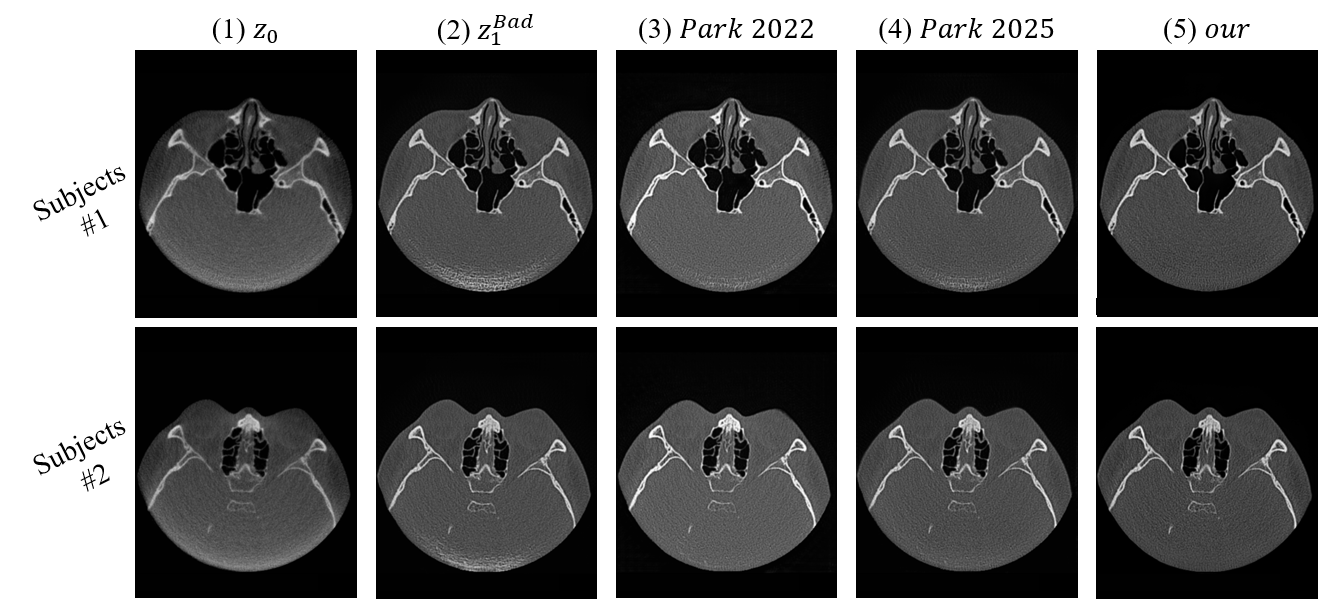}
	\caption{
		Qualitative comparison of high-LPIPS examples from the \(\bm{z}_1^{\text{bad}}\) set. Each row illustrates a representative axial slice from a different subject affected by shade artifacts. 
		From left to right: (1) input CBCT image \(\bm{z}_0\), (2) pseudo-target \(\bm{z}_1^{\text{bad}}\) generated by the pretrained CycleGAN~\cite{park2022unpaired}, (3) enhanced output from Park2022~\cite{park2022unpaired}, trained in a paired setting, (4) fine-tuned version from Park2025~\cite{park2025shade}, and (5) output from our proposed Schrödinger Bridge-based method.
	}
	\label{fig:lpips-bad-examples}
\end{figure}

To illustrate this contrast, we selected examples with the highest LPIPS scores from the \(\bm{z}_1^{\text{bad}}\) set and visualize them in Figure~\ref{fig:lpips-bad-examples}. As shown, the baseline images generated by \(G_s\) exhibit severe shade artifacts, particularly in soft-tissue and posterior cranial regions. Although these SB-based outputs yield high LPIPS scores—indicating low similarity to artifact-laden baselines—they show clearer anatomical delineation and notably suppressed artifacts. These findings reveal that LPIPS is sensitive enough to capture meaningful perceptual differences caused by artifact suppression, making it a valuable tool for interpreting improvements in clinically compromised inputs.

\begin{table}[t]
	\centering
	\caption{ARR and ARSR comparison for shade artifact suppression on \(\bm{z}_{\mathrm{SB}}^{\text{bad}}\) outputs across training and test datasets. Arrows indicate preferred direction: ARR and ARSR $\uparrow$(higher is better).}
	\label{tab:arr-arsr-results}
	\resizebox{\columnwidth}{!}{
		\begin{tabular}{lcccc}
			\toprule
			\textbf{Method} & \textbf{ARR $\uparrow$ (Train)} & \textbf{ARSR $\uparrow$ (Train)} & \textbf{ARR $\uparrow$ (Test)} & \textbf{ARSR $\uparrow$ (Test)} \\
			\midrule
			\textbf{G\textsubscript{SB} (Ours)} & \textbf{98.25} & \textbf{99.35} & \textbf{96.98} & \textbf{96.23} \\
			Park2025~\cite{park2025shade}      & 97.75 & 99.30 & 96.86 & 95.96 \\
			Park2022~\cite{park2022unpaired}   & 97.58 & 98.74 & 95.74 & 95.96 \\
			\bottomrule
		\end{tabular}
	}
\end{table}

\subsection{Fidelity and Structural Similarity}
To assess structural fidelity beyond artifact suppression, we compared \(\bm{z}_1^{\text{good}}\), produced by the pretrained CycleGAN generator \(G_s\), with \(\bm{z}_{\mathrm{SB}}^{\text{good}}\), generated by our proposed model. In line with the prior protocol~\cite{park2022unpaired}, we computed DC, RMSE, and SSIM with respect to the expert-approved reference images.

In addition to these conventional metrics, we employed the LPIPS~\cite{zhang2018unreasonable} to better capture perceptual discrepancies between generated images and the references. Unlike RMSE and SSIM, which focus on pixel-wise or luminance-based comparisons, LPIPS operates in a deep feature space derived from pretrained neural networks, making it more aligned with human perceptual judgments. This allows for finer discrimination of structural inconsistencies and subtle artifacts that may be clinically relevant yet challenging to quantify using traditional measures.

The results are summarized in Table~\ref{tab:quantitative-results}. Our method (\(G_{\mathrm{SB}}\)) achieves the best performance across all evaluation metrics, including the lowest RMSE and LPIPS, as well as the highest SSIM and DC. Notably, \(G_{\mathrm{SB}}\) outperforms the fine-tuned Park2025~\cite{park2025shade} baseline by a significant margin in perceptual similarity (LPIPS: 0.0015 vs. 0.0108) and accuracy of structural correspondence (RMSE: 0.0030 vs. 0.0081). Compared to the earlier Park2022~\cite{park2022unpaired} model, our approach demonstrates an even more pronounced improvement, particularly in SSIM (0.9971 vs. 0.9130) and LPIPS (0.0015 vs. 0.0333). These results confirm that our Schrödinger Bridge framework, enhanced with human-guided conditional diffusion, more effectively preserves anatomical fidelity while suppressing clinically undesirable artifacts.

\begin{table}[t]
	\centering
	\caption{Fidelity and structural similarity results for $\bm{z}_\mathrm{SB}^\text{good}$ on train and test sets. RMSE, SSIM, LPIPS, and DC are reported as mean $\pm$ standard deviation.}
	\label{tab:quantitative-results}
	\resizebox{\columnwidth}{!}{
		\begin{tabular}{lcccc}
			\toprule			
			\textbf{Method(Train)} & \textbf{RMSE $\downarrow$} & \textbf{SSIM $\uparrow$} & \textbf{LPIPS $\downarrow$} & \textbf{DC (\%) $\uparrow$} \\

			\midrule
			\textbf{G\textsubscript{SB} (Ours)} & $0.0025 \pm 0.0007$ & $0.9975\pm 0.0010$ & $0.0011\pm 0.0006$ & $81.61 \pm 6.29$ \\
			Park2025~\cite{park2025shade} & $0.0085 \pm 0.0025$ & $0.9895\pm 0.0031$ & $0.0113\pm 0.0033$ & $80.60 \pm 6.73$ \\
			Park2022~\cite{park2022unpaired} & $0.0259 \pm 0.0062$ & $0.9081\pm 0.0160$ & $0.0375\pm 0.0105$ & $81.35 \pm 6.93$ \\

			\toprule
			\textbf{Method(Test)} & \textbf{RMSE $\downarrow$} & \textbf{SSIM $\uparrow$} & \textbf{LPIPS $\downarrow$} & \textbf{DC (\%) $\uparrow$} \\

			\midrule
			\textbf{G\textsubscript{SB} (Ours)} & $0.0030 \pm 0.0010$ & $0.9971 \pm 0.0013$ & $0.0015 \pm 0.0009$ & $83.95 \pm 4.97$ \\
			Park2025~\cite{park2025shade} & $0.0081 \pm 0.0022$ & $0.9898 \pm 0.0029$ & $0.0108 \pm 0.0034$ & $82.60 \pm 5.67$ \\
			Park2022~\cite{park2022unpaired} & $0.0235 \pm 0.0053$ & $0.9130 \pm 0.0100$ & $0.0333 \pm 0.0105$ & $83.84 \pm 5.66$ \\

			\bottomrule
		\end{tabular}
	}
\end{table}

\begin{table}[t]
	\centering
	\caption{
		Quantitative performance of the proposed SB-based model on the $\bm{z}_\mathrm{SB}^{\text{good}}$ test set under different NFE. All metrics (RMSE, SSIM, and LPIPS) are computed with respect to the pseudo-target $\bm{z}_1^{\text{good}}$ generated by the pretrained GAN-based model. Although fidelity slightly decreases as NFE increases, the variation remains within an acceptable range for clinical application, supporting the robustness of our method even with fast sampling.
	}	
	\label{tab:sampling-efficiency}
	\resizebox{\columnwidth}{!}{
		\begin{tabular}{c|c|c|c}
			\toprule
			\textbf{NFE} & \textbf{RMSE $\downarrow$} & \textbf{SSIM $\uparrow$} & \textbf{LPIPS $\downarrow$} \\
			\midrule
			10     & $0.0030 \pm 0.0010$ & $0.9971 \pm 0.0013$ & $0.0015 \pm 0.0009$ \\
			100    & $0.0035 \pm 0.0010$ & $0.9953 \pm 0.0014$ & $0.0018 \pm 0.0009$ \\
			1000  & $0.0048 \pm 0.0009$ & $0.9868 \pm 0.0017$ & $0.0053 \pm 0.0012$ \\		
			\bottomrule
		\end{tabular}
	}
\end{table}

\subsection{Sampling Efficiency}
To better understand the trade-off between sampling cost and output quality, we investigated how the number of sampling steps affects generation in our SB-based model. Unlike conventional diffusion models such as DDPM~\cite{ho2020denoising} or SR3~\cite{saharia2022sr3}, which typically require hundreds or thousands of steps starting from random noise, our approach leverages a stochastic bridge between two real boundary distributions, thereby enabling more efficient sampling.

Following the findings of I\textsuperscript{2}SB~\cite{liu20232}, which demonstrated that high-quality samples can be produced with a small number of steps, we set the number of function evaluations (NFE) to 10 in our framework. Consistent with their findings, our application to medical image translation confirms that even with only 10 sampling steps, the model preserves both anatomical structure and perceptual quality. 

\begin{figure}[t]
	\centering

	\includegraphics[width=\textwidth]{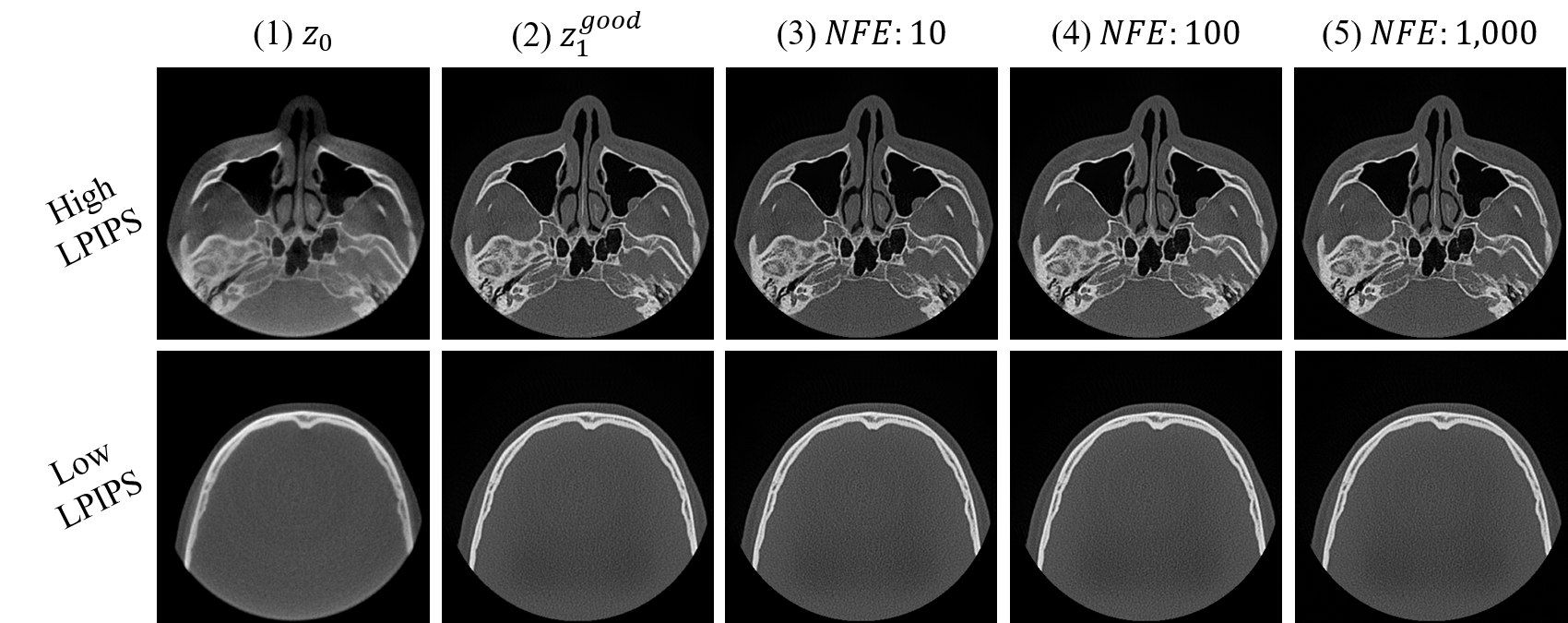}
	\caption{
		Visual comparison of SB-generated outputs under different NFE. Examples are selected from the $\bm{z}_\mathrm{SB}^\text{good}$ test set and correspond to the lowest and highest LPIPS scores at NFE = 10, allowing us to visualize the best- and worst-case scenarios in fast sampling. Despite the drastic reduction in iteration count, low-NFE outputs preserve both anatomical structure and artifact suppression, validating the quantitative results in Table~\ref{tab:sampling-efficiency} and highlighting the clinical viability of accelerated generation.
	}	
	\label{fig:sampling-efficiency-vis}
\end{figure}

To quantitatively assess this, we evaluated RMSE, SSIM, and LPIPS across varying NFE configurations using expert-approved reference images. Specifically, we followed the same evaluation procedure described in the \textit{Fidelity and Structural Similarity} section. Table~\ref{tab:sampling-efficiency} summarizes the results. A visual comparison across different NFE is provided in Figure~\ref{fig:sampling-efficiency-vis}.

Interestingly, we observed a counterintuitive decline in fidelity metrics (e.g., LPIPS, SSIM) with increasing NFE. This effect is especially evident when measuring against pseudo-references generated by a pretrained GAN~\cite{park2022unpaired}, which may contain shade artifacts or structural biases. We attribute this trend to two factors: (1) over-smoothing due to excessive sampling, which blurs fine anatomical details~\cite{xia2024timetuner,liu20232}; and (2) distributional drift away from the artifact-prone reference, effectively reflecting improvements not captured by standard perceptual metrics.

This phenomenon aligns with the observations by Xia et al.~\cite{xia2024timetuner}, who report that excessive sampling in fast samplers can introduce truncation errors, leading to distributional drift and quality degradation. In our case, we hypothesize that extended sampling introduces over-smoothing, which may blur fine anatomical features and reduce perceptual similarity. Nevertheless, the performance gap across different sampling steps remained within an acceptable range in qualitative assessments, suggesting that the observed metric variations are largely numerical artifacts induced by accumulated truncation error, rather than reflecting meaningful perceptual degradation. 

These findings highlight the practicality of our SB-based method, particularly in clinical scenarios where low-latency image generation is critical. Notably, expert radiologists confirmed that the outputs—even at higher sampling steps—remained within clinically acceptable limits, suggesting that the observed degradation in quantitative metrics does not necessarily reflect a perceptual decline.

\begin{figure}[t]
	\centering

	\includegraphics[width=\textwidth]{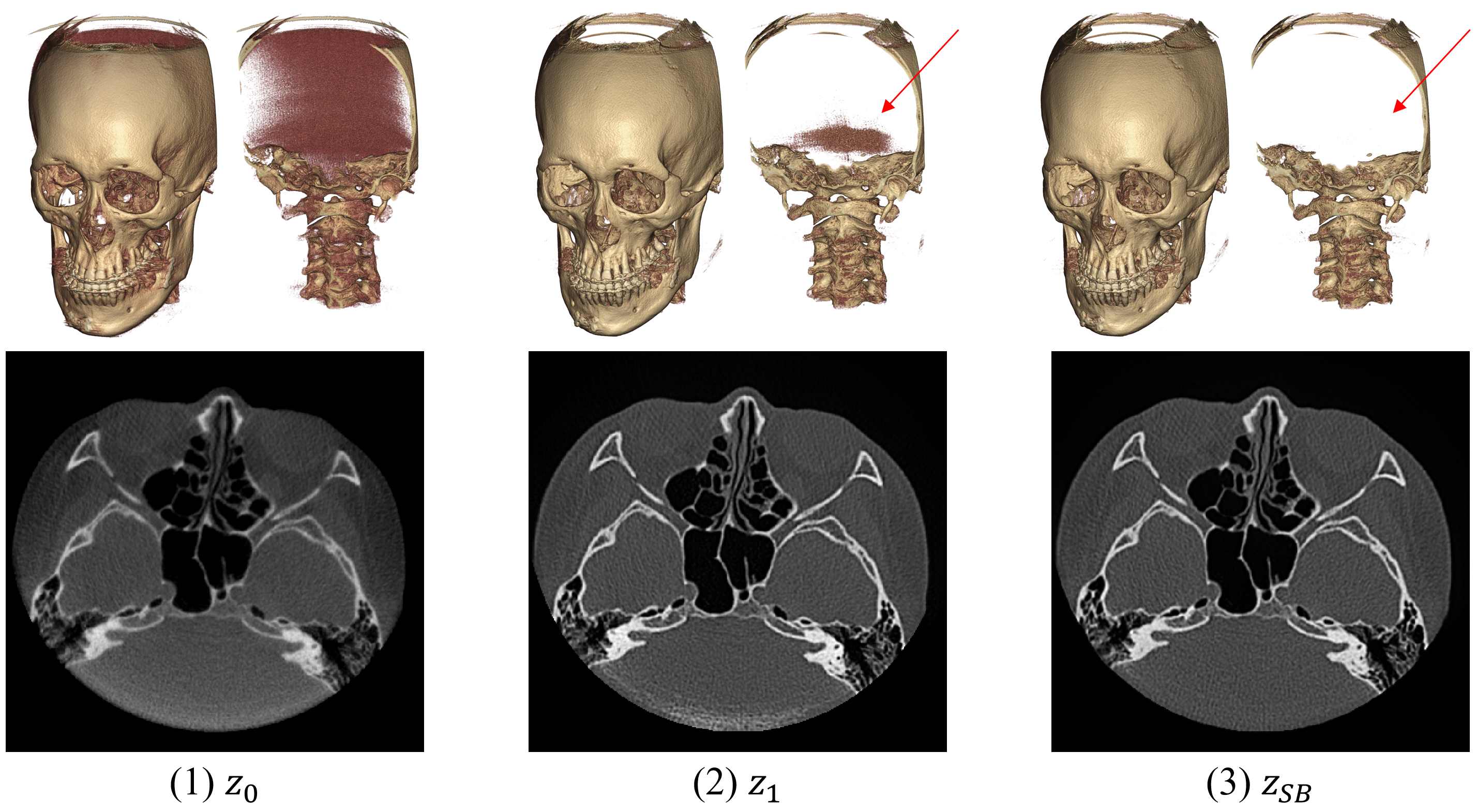}
	\caption{
		Volume rendering comparison from the same subject under different generation models. From left to right: (1) volume rendering of the input CBCT image, (2) volume rendering of the image generated by the pretrained model \(G_s\), and (3) volume rendering of the image generated by the proposed method. For each case, the entire cranial volume and the occipital region—identified as a common site of shade artifacts—were rendered using the same bone-specific transfer function, with key abnormalities indicated by red arrows for visual reference. The bottom row shows the corresponding axial slices. Compared to the baseline outputs, the proposed method significantly suppresses shade artifacts in the posterior region while preserving anatomical details throughout the volume.
	}
	\label{fig:volume_rendering}
\end{figure}

\subsection{Volume-Rendered Qualitative Evaluation}

To further assess the clinical relevance of our method, we performed volume rendering of the full CBCT volumes to examine global anatomical consistency and the distribution of shade artifacts across all slices. As shown in Figure~\ref{fig:volume_rendering}, the entire skull and occipital region were visualized using the same bone-specific transfer function. The leftmost column shows the input CBCT image, which often contains severe shade artifacts. The center and right columns show outputs from the pretrained model $G_s$ and our proposed method, respectively. 

In addition to full-volume rendering, we also performed ROI cropping around the occipital lobe—a region highly susceptible to shading—to better visualize artifact suppression. The bottom row shows the corresponding axial slices for further comparison. The volume rendering clearly demonstrates that our model significantly reduces high-intensity shade artifacts—especially in the posterior region—while preserving anatomical details throughout the volume. These findings confirm that our approach not only preserves anatomical integrity but also effectively suppresses non-structural artifacts in a volumetric context. Such volume-based visualization reflects practical diagnostic workflows, where consistent artifact suppression across slices is critical for surgical planning and interpretation.

All volume-rendered visualizations, including full skull views and ROI crops, were visualized using 3D Slicer~\cite{fedorov20123dslicer}, an open-source medical imaging platform widely used in clinical research.

\subsection{Ablation of conditioning CBCT image}
\begin{figure}[t]
	\centering
	\includegraphics[width=\textwidth]{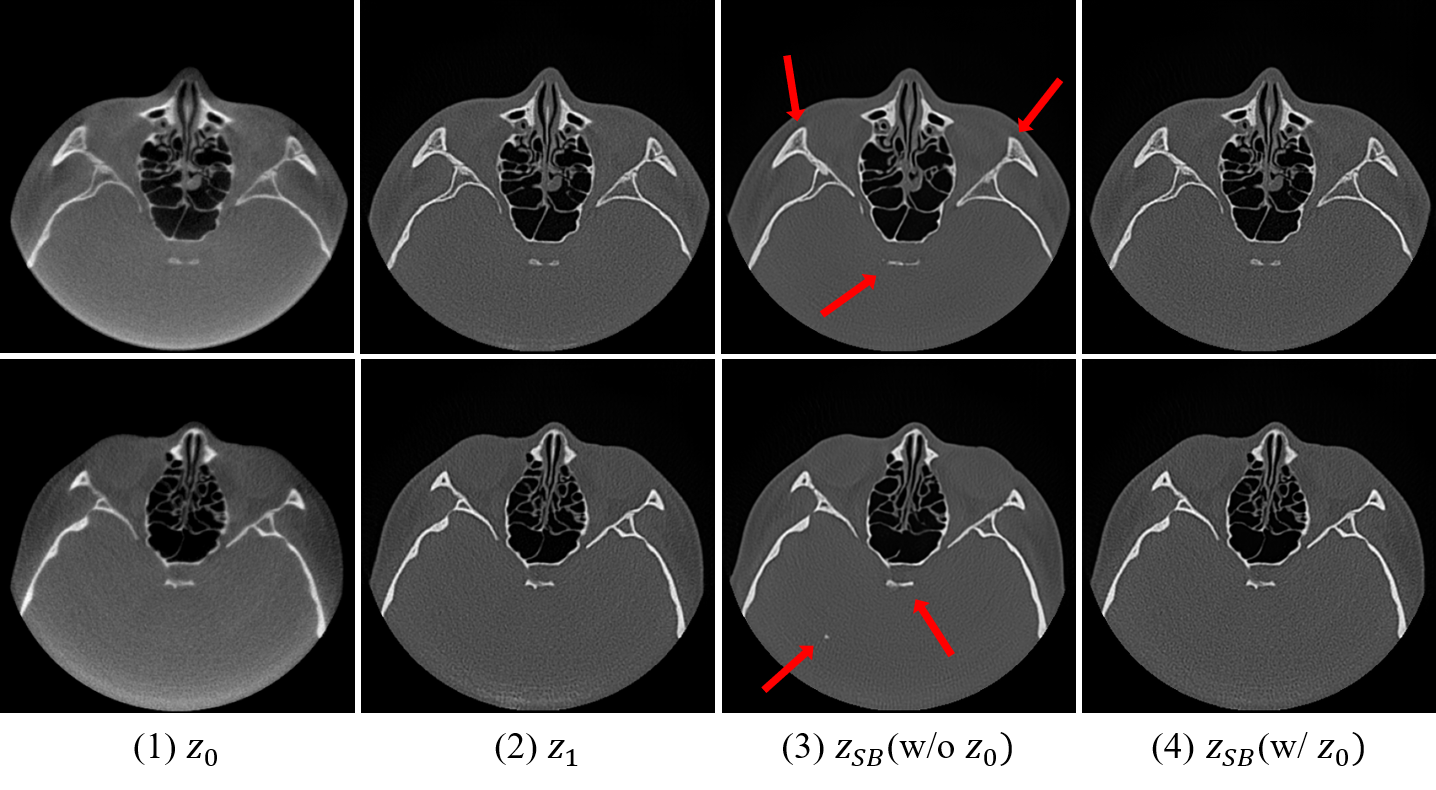}
	\caption{
		Ablation of conditioning on \(\bm{z}_0\). From left to right: (1) input CBCT image \(\bm{z}_0\), (2) pseudo-target \(\bm{z}_1\) generated by the pretrained GAN-based generator, (3) output generated without conditioning on \(\bm{z}_0\), and (4) output generated with full conditioning. The third image reveals anatomical inconsistencies and texture mismatches, as highlighted by red arrows. These results confirm the necessity of prior conditioning to preserve anatomical fidelity.
	}
	\label{fig:result_without_z0_conditioning}
\end{figure}

To investigate the role of the input CBCT image $\bm{z}_0$ in both training and sampling stages, we conducted an ablation study by removing $\bm{z}_0$ from the conditioning path. Our goal was to assess how the absence of this spatial prior affects the anatomical fidelity of the generated outputs. As shown in Figure~\ref{fig:result_without_z0_conditioning}, the resulting images failed to preserve anatomical plausibility in key regions. While the generated image may appear realistic at a glance, closer inspection—highlighted by red arrows—reveals distortions and deviations from the expected structural layout. Furthermore, the generated texture clearly differs from that of the pseudo-target $\bm{z}_1$, indicating poor alignment with the intended style and content.

Quantitatively, the degradation is also evident. On the held-out test set, the model without $\bm{z}_0$ conditioning yielded an RMSE of $0.0224 \pm 0.0060$, SSIM of $0.8510 \pm 0.0289$, and LPIPS of $0.0825 \pm 0.0144$, all of which reflect a significant drop in perceptual and structural quality compared to the fully conditioned model. These findings confirm that $\bm{z}_0$ plays a crucial role as a spatial and anatomical anchor in both learning and sampling. Its absence impairs the model’s ability to reconstruct clinically faithful images (cf.~Table~\ref{tab:quantitative-results}).

\subsection{Negative Preference Request}
While the primary objective of our framework is to reduce shade artifacts by guiding generation toward the “good” class, we further investigate the model’s response to inverted preference signals. Specifically, we analyze whether the model can generate controlled outputs that exhibit shade artifacts when explicitly guided to do so. This “negative request” setting involves conditioning on inputs from \(\bm{z}_0^{\text{good}}\)—CBCT images that originally produce artifact-free outputs—while providing a negative preference request as conditioning signal.

\begin{figure}[t]
	\centering
	\includegraphics[width=\linewidth]{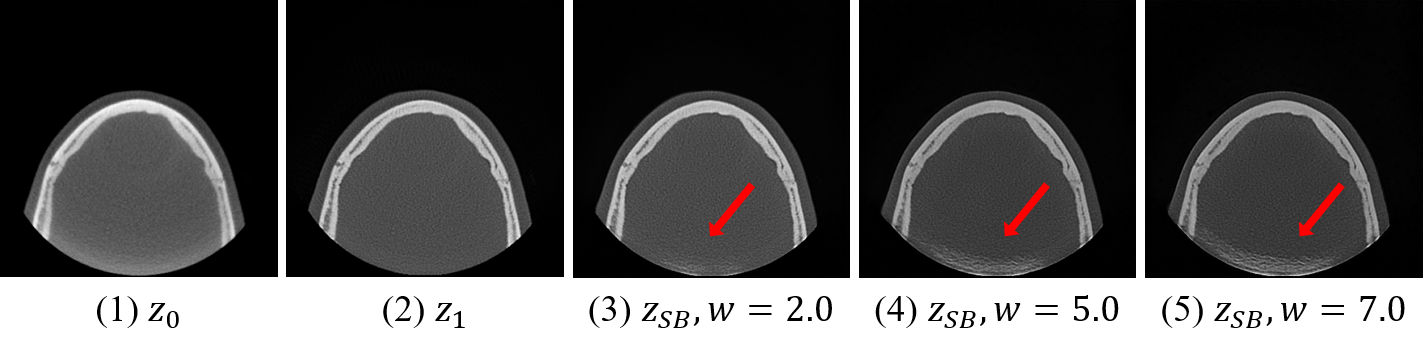}
	\caption{
		Visualization of outputs under negative preference request. From left to right: (1) input artifact-free CBCT image \(\bm{z}_0^{\text{good}}\) from the test set, (2) pseudo-target \(\bm{z}_1^{\text{good}}\) generated by the pretrained GAN-based generator, and (3–5) outputs generated by the model under “bad” preference conditioning with increasing guidance strengths \(w\). The generated artifacts, particularly in the occipital region (red arrows), closely resemble those typically observed in real artifact cases. These results demonstrate the model's ability to control artifact presence via preference conditioning, confirming its semantic controllability and interpretability.
	}
	\label{fig:negative-request}
\end{figure}

Interestingly, due to the inherent stochasticity and generative diversity of diffusion-based models, we observe that our method is capable of synthesizing plausible shade artifacts even in slices where such artifacts did not originally exist(test set). As illustrated in Figure~\ref{fig:negative-request}, the model responds to “bad” preference conditioning by introducing shade artifacts, particularly in soft-tissue regions such as the occipital area, mimicking the artifact patterns typically seen in real \(\bm{z}_1^{\text{bad}}\) cases.

This behavior demonstrates that the model has not merely memorized training artifacts but has learned an interpretable semantic space wherein artifact presence can be modulated through binary feedback. Such controllability underscores the explainable nature of our framework, offering potential for both artifact suppression and controlled artifact simulation—useful, for instance, in data augmentation or robustness testing scenarios.

Although our current task focuses on a unidirectional preference-guided generation setting—specifically, transforming “bad” images into “good” ones—this capability implies broader applicability. The inherent diversity of diffusion-based models suggests that future extensions could incorporate bidirectional or even multi-class preference guidance. This opens the door to more fine-grained, multi-attribute conditional sampling, where generation can be steered across a spectrum of semantic qualities using scalar or categorical feedback signals.

\section{Conclusion}

We introduced a diffusion-based framework for CBCT-to-MDCT translation that leverages Schrödinger Bridge dynamics, classifier-free guidance, and human preference feedback. By explicitly modeling boundary states \(\bm{z}_0\) and \(\bm{z}_1\), and conditioning generation on binary feedback signals, the proposed method effectively reduces shade artifacts while maintaining anatomical fidelity.

Our results show that the SB formulation improves sample diversity and interpretability compared to traditional diffusion or GAN-based approaches. Conditioning on \(\bm{z}_0\) is essential for spatial consistency, while preference-guided CFG sampling enables controllable artifact suppression and refinement. The model supports rapid sampling with only 10 steps, significantly reducing inference time without compromising quality.

Beyond artifact reduction, we demonstrate that our model can respond to both positive and negative preference inputs, revealing a controllable latent space aligned with clinical semantics. Volume-rendered evaluations and ablation studies further validate the model’s performance and explainability.

Together, these contributions establish a practical foundation for user-aligned, feedback-driven generation in medical imaging. Our framework is scalable, interpretable, and adaptable to real-world clinical use, with potential applications in personalized image enhancement, artifact-aware simulation, and diagnostic support across imaging modalities.

\vspace{1mm}
\noindent\textbf{Acknowledgement.} This work was supported in part by the National Research Foundation of Korea (NRF) Grant funded by Korean Government [Ministry of Science and ICT (MSIT)] under Grant RS-2024-00338504, and in part by the National Institute for Mathematical Sciences (NIMS) funded by Korean Government under Grant NIMS-B25910000.

{\small
	\bibliographystyle{unsrtnat}
	\bibliography{ref}
}

\end{document}